\documentclass[pdflatex,sn-mathphys-num]{sn-jnl}

\usepackage{graphicx}%
\usepackage{multirow}%
\usepackage{amsmath,amssymb,amsfonts}%
\usepackage{amsthm}%
\usepackage{mathrsfs}%
\usepackage[title]{appendix}%
\usepackage{xcolor}%
\usepackage{textcomp}%
\usepackage{manyfoot}%
\usepackage{booktabs}%
\usepackage{makecell}
\usepackage{multirow}
\usepackage{algorithm}%
\usepackage{algorithmicx}%
\usepackage[noend]{algpseudocode}
\usepackage{listings}%

\usepackage{forest}
\usepackage{xcolor}

\definecolor{foldercolor}{RGB}{100, 149, 237}
\definecolor{rootcolor}{RGB}{70, 130, 180}

\usepackage{algorithm}
\usepackage{graphicx}
\usepackage{booktabs}
\usepackage{longtable}
\usepackage{tabularx}
\usepackage{pifont}

\usepackage[capitalize]{cleveref}
\usepackage[x11names]{xcolor}
\usepackage{subcaption}
\usepackage{dirtree}
\usepackage{caption}
\crefformat{figure}{#2figure~#1#3}
\Crefformat{figure}{#2Figure~#1#3}

\theoremstyle{thmstyleone}%
\theoremstyle{thmstyletwo}%

\theoremstyle{thmstylethree}%

\begin{document}

\title[ROOFS: RObust biOmarker Feature Selection]{ROOFS: RObust biOmarker Feature Selection}

\author*[1]{\fnm{Anastasiia} \sur{Bakhmach}}\email{anastasiia.bakhmach@inria.fr}
\author[1]{\fnm{Paul} \sur{Dufossé}}
\author[1]{\fnm{Simon} \sur{Charpigny}}
\author[2]{\fnm{Florence} \sur{Monville}}
\author[1,3]{\fnm{Laurent} \sur{Greillier}}
\author[4]{\fnm{Fabrice} \sur{Barlési}}
\author[1]{\fnm{Sébastien} \sur{Benzekry}}

\affil*[1]{\orgdiv{COMPO Team (COMPutational Pharmacology and clinical Oncology)}, \orgname{Inria Sophia Antipolis -- Méditerranée, Inserm U1068, CNRS UMR7258, Institut Paoli-Calmettes, Aix-Marseille University}, \orgaddress{\city{Marseille}, \country{France}}}
\affil[2]{\orgname{Veracyte SAS}, \orgaddress{\city{Marseille}, \country{France}}}
\affil[3]{\orgname{Assistance Publique-Hôpitaux de Marseille (APHM)}, \orgaddress{\city{Marseille}, \country{France}}}
\affil[4]{\orgname{Gustave Roussy}, \orgaddress{\city{Marseille}, \country{France}}}




\abstract{\textbf{Background:} Feature selection (FS) is essential for biomarker discovery and clinical predictive modeling. Over the past decades, the methodological literature on FS has become rich and mature, offering a wide spectrum of algorithmic approaches. However, much of this methodological progress has not been fully adopted in applied biomedical research.

\textbf{Results:} To help bridge this gap, we propose \textit{ROOFS} (RObust biOmarker Feature Selection), a Python package designed to help researchers in the choice of the FS method most adapted to their problem. \textit{ROOFS} benchmarks multiple FS methods on the user's data and generates reports summarizing a comprehensive set of evaluation metrics, including downstream predictive performance estimated using optimism correction, stability, robustness of individual features, and true positive and false positive rates assessed on semi-synthetic data with a simulated outcome. We demonstrate the utility of \textit{ROOFS} on data from the PIONeeR biomarkers study, aimed at identifying predictors of resistance to anti-PD-(L)1 immunotherapy in lung cancer. 

\textbf{Conclusion:} Comprehensive benchmarking with \textit{ROOFS} has the potential to improve the reproducibility of FS discoveries and increase the translational value of clinical models.

\textbf{Availability and implementation:} ROOFS is available at \url{https://gitlab.inria.fr/compo/roofs}.
}



\maketitle

\section{Introduction}

Feature selection (FS) comprises a class of methods aimed at reducing the feature space by selecting a subset of size $k < p$, where $p$ denotes the total number of features. In supervised learning, FS is used either for discovery, to identify the most informative variables, or for prediction, to reduce variance. FS methods can be classified into 4 groups: filter, embedded, wrapper, and ensemble methods (see \cite{Guyon_Elisseeff_2003} and \cite{data_perspective} for extensive reviews). Filters evaluate features according to specific criteria and output either a complete ranking or a subset selected using a predefined threshold (e.g., a statistical test). Embedded methods perform FS simultaneously with model learning, as in regularization-based approaches or tree-based models with constrained depth. Wrapper methods iteratively search for feature subsets that maximize the predictive performance of a given model. Ensemble methods aggregate outputs from different selection algorithms or from repeated runs of a single method on resampled data. Filter and embedded methods are typically more straightforward, computationally efficient, and interpretable compared to wrapper and ensemble techniques. Computationally expensive and algorithmically complex wrapper methods are capable of capturing feature interactions tailored to a specific classifier, typically resulting in modest performance improvement. The computational cost of the ensemble methods is offset by their ability to select features with the strongest signal and avoid false discoveries.

A basic requirement for FS is that the reduced model should perform equivalently to or better than the full model, capturing the complexity of the original data. In the context of biomarker discovery and clinical predictive models, the selected subset, often called a ``signature'', should meet additional requirements: (i) contain biologically or clinically interpretable features; (ii) be as minimal as possible; and (iii) demonstrate \textit{stability}, i.e. the selected features should not be sensitive to small changes in the data \cite{Kalousis_2005}. FS stability is of key importance for ensuring reproducibility. It is typically assessed by measuring the similarity between feature subsets obtained across resamples of the original data, e.g. via bootstrapping or subsampling. An unstable FS method produces feature subsets that change substantially across resamples, so the discovered biomarkers are unreliable and may not replicate in external data. In addition to compromising biomarker validity, this instability may result in models that generalize poorly and are therefore unlikely to be clinically useful.

Feature selection is a long-standing field of research in machine learning \cite{Guyon_Elisseeff_2003}. However, much of the methodological progress has not fully translated into applied biomedical research, where practitioners often rely on at most a limited set of familiar methods (e.g. LASSO and Boruta). Likewise, bootstrap-based procedures for assessing stability are rarely used, despite its importance having been frequently acknowledged \cite{Kalousis_2005, Haury_Gestraud_Vert_2011, Heinze_2018}. In addition, the suitability of FS methods for a given dataset is rarely evaluated systematically, although previous benchmark studies have shown that, in terms of predictive performances, the optimal method is contingent upon the specific dataset \cite{BOMMERT2020106839, Demircioglu_2025}. Furthermore, FS performance is sensitive to hyperparameter choices, such as the selected subset size, and different settings can lead to different rankings of methods on the same dataset \cite{Tatwani2019SubsetSize}.

To address this gap between methodological and applied research, we propose \textit{ROOFS} (RObust biOmarker Feature Selection), a Python package available at \underline{https://gitlab.inria.fr/compo/\textit{roofs}}. It implements a pipeline that, in a single run on a user-provided dataset, compares multiple models, enables identification of the optimal model, and facilitates data exploration by measuring feature robustness. \textit{ROOFS} conducts comprehensive benchmarking of 45 pre-implemented FS algorithms on the user's data based on three criteria: 1) method stability, 2) predictive performance, and 3) optionally, true positive and false positive rates (from semi-synthetic data with a simulated outcome). For point 2), we relied on the optimism-correction (OC) framework, emphasized as a guideline for the development of clinical predictive models by \citet{Collinse074819}.

\citet{Balikci_Njume_Cakmak_2026} provide a review of similar tools designed for the purpose of biomarker discovery. The key contribution of \textit{ROOFS} that distinguishes it from existing approaches is its emphasis on stability and bootstrap-based internal validation with optimism correction. 

We demonstrate the practical value of \textit{ROOFS} by applying it to a complex dataset from the PIONeeR biomarkers clinical study (NCT03833440, Precision Immuno-Oncology for Advanced Non-small Cell Lung Cancer Patients With PD-1 ICI Resistance) to predict primary resistance to immunotherapy. This dataset is an interesting and challenging use case for FS because: 1) the data is rich and heterogeneous combining clinical features, routine blood tests and detailed immunoprofiling of blood and tumor samples measured by different clinical and research partners; 2) it contains clusters of highly correlated features, e.g. numerous variables describing related immune cell populations; and 3) $p$ (374) is roughly equivalent to $n$ (435). This is different from the three types of datasets popular in the literature: a) high-dimensional omics ($p \gg n$); b) data with a sufficiently high event-per-variable ratio, e.g. the number of patients in the smallest class is at least ten times the number of predictors; and c) big data (e.g., from electronic health records), with $n \gg p$ typically modeled using deep artificial neural networks. 

\section{Implementation}

This section is organized as follows. Subsection \ref{subsec:package-structure} provides an overview of the main components of \textit{ROOFS}. Subsection \ref{subsec:methods} describes in detail the methodology implemented in the package to evaluate FS methods. Subsection \ref{subsec:highlighted-functionalities} highlights additional features of the package that may be of particular interest to users.

\subsection{Package overview}
\label{subsec:package-structure}

\textit{ROOFS} is composed of several modules: data preprocessing, feature selection, performance estimation via optimism correction, synthetic outcome generation, and auxiliary functions. The first three modules are combined into a benchmarking pipeline; the functions implementing the FS methods can also be used independently of this pipeline. The package is available in two versions. The core version provides classical FS methods, such as LASSO and t-test p-value filtering, with minimal dependencies for quick installation. The full version offers 45 FS methods, including more advanced algorithms, and is better suited for comprehensive benchmarking; it requires several additional dependencies to be installed.

\subsubsection{Data}
\begin{figure}[ht]
    \centering
    \includegraphics[
        trim=0cm 0cm 0cm 6.2cm,
        clip,
        width=0.95\textwidth
    ]{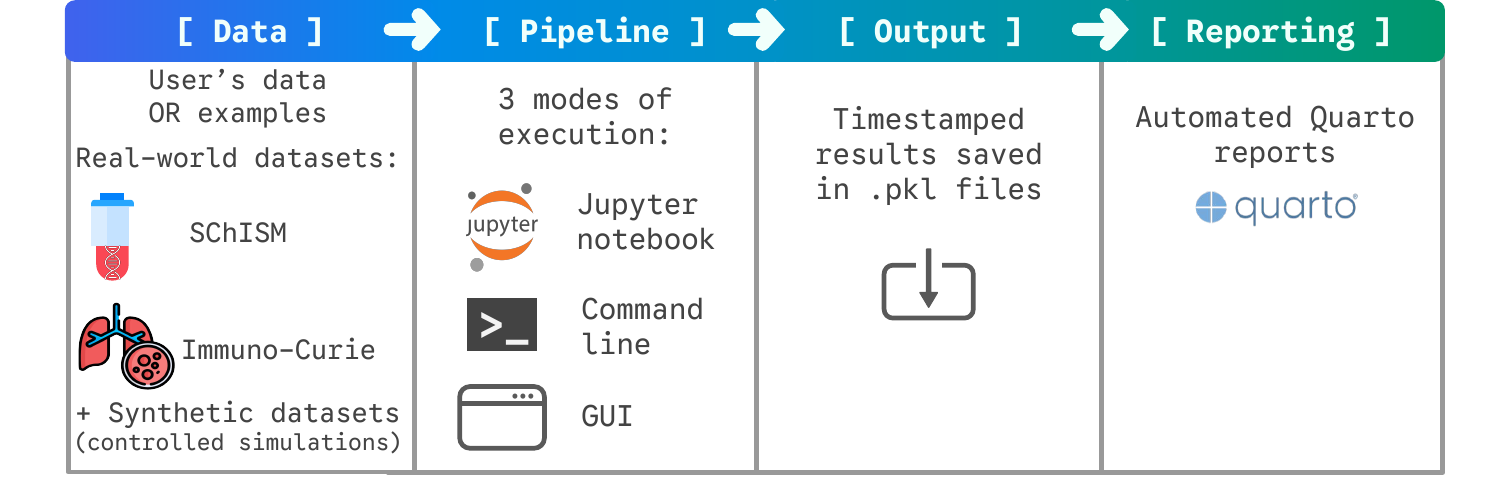}
    \caption{Overview of \textit{ROOFS} workflow. The pipeline accepts user-provided or built-in real-world (SChISM, Immuno-Curie) or synthetic datasets. Three execution modes are supported: Jupyter notebook, command line, and graphical user interface (GUI). Results are saved as timestamped .pkl files for traceability, and automated Quarto reports are generated with a summary of key performance metrics.}
    \label{fig:structure}
\end{figure}

The workflow for benchmarking FS methods with \textit{ROOFS} is illustrated in Fig. \ref{fig:structure}.  The user-provided data is taken as an input. Alternatively, two publicly available datasets are provided in the package for tutorial and testing. The SChISM (Size CfDNA Immunotherapy Signature Monitoring) dataset includes pre-treatment plasma cfDNA size profiles and clinical variables (150 features total) from 126 advanced and/or metastatic cancer patients treated with immune checkpoint inhibitors (ICI) as monotherapy or in combination with chemotherapy or targeted therapy. The prediction task is to evaluate whether size-based cfDNA concentration can serve as a biomarker for immunotherapy response \cite{schism_zenodo, schism_preprint}. The Immuno-Curie dataset contains 198 radiomic, pathomic, and transcriptomic features collected for 317 metastatic non-small cell lung cancer (NSCLC) patients to predict immunotherapy outcomes \cite{curie_zenodo, curie_dataset}.

\subsubsection{Pipeline execution}

Three approaches for executing a benchmarking pipeline are provided (Fig. \ref{fig:structure}): a) a regular Jupyter notebook suitable for exploratory use, b) a command-line interface, which allows linking multiple pipelines within a Bash script or integrating \textit{ROOFS} with other tools, and c) a Streamlit-based graphical user interface (GUI) for users with limited programming experience (Supplementary fig. S1). Upon pipeline execution, model performance metrics and selected features are saved in a .pkl file, which may then be used for automated report generation.

\subsubsection{Automated reporting}

To allow users to assess the results of the benchmarking, \textit{ROOFS} generates a report with a summary of key performance metrics. These reports include: 1) a table that aggregates FS frequencies across methods and shows agreement between different methods in selecting individual features; 2) a stability – AUC plot for a quick visual overview; 3) an interactive customizable bar plot visualizing performance metrics of different models, accompanied by a summary table with filtering options that help identify the optimal methods based on the user's criteria of interest. For an example, see \url{https://roofs-ce5754.gitlabpages.inria.fr/index.html}.

\subsubsection{Choice of an optimal FS method}

The package does not provide an optimal FS method for a given dataset, because the appropriate trade-off between evaluation metrics should be guided by the specific predictive or discovery goals. Relevant questions include: Is a sparser signature preferred over a more comprehensive set of biomarkers? Which performance metric should be prioritized (e.g., positive predictive value or sensitivity)? Does a small gain in performance outweigh improvements in stability?

\subsection{Methods}
\label{subsec:methods}
The evaluation of a given FS method by \textit{ROOFS} is based on a bootstrap experiment (Fig. \ref{fig:scheme}A). Let $ \mathcal{D}_{\text{original}} = \{(x^i, y^i)\}_{i=1}^{n} $ denote the original dataset provided by the user. From this, $B$ bootstrap resamples are derived, each denoted $\mathcal{D}_b$, $b \in [1, B]$. An identical pipeline, including data preprocessing (imputation using the median for continuous variables or mode for categorical variables and z-score normalization), FS and subsequent predictive modeling, is applied to both the full data and each bootstrap, and repeated across all user-specified combinations of FS methods and classifiers.

\subsubsection{FS methods}
\textit{ROOFS} gathers 45 FS methods from all algorithmic families – 21 filters, 10 embedded methods, 7 wrappers, and 7 ensembles – including both classical statistical techniques and recent ML-based approaches. Most implementations were adapted from existing Python libraries, including scikit-learn \cite{scikit-learn}, scikit-feature \cite{data_perspective}, Stabl \cite{Hedou_Maric_2024}, and others. 

FS methods gathered in \textit{ROOFS} can be divided into two groups based on how they determine the number of features they select: 1) 19 methods with fixed, user-defined subset size, and 2) 26 methods with varying subset size dependent on internal criteria, such as cross-validated lambda in the case of LASSO or p-value thresholds in the case of p.adjust.

Among the 45 FS methods implemented in \textit{ROOFS}, a representative subset of 23 approaches covering all four methodological families was chosen for benchmarking on the PIONeeR dataset. Among embedded methods, we evaluated LASSO \cite{Tibshirani_1996} and its variants (adaptive LASSO \cite{Zou_2006} and exclusive LASSO \cite{Zhou_Jin_Hoi_2010}), as well as ensemble bootstrap-based methods designed to enhance stability and control false discoveries (modified Bolasso \cite{Bach_2008} with a frequency threshold of 0.5 followed by a second LASSO fit, stability selection \cite{Meinshausen_Buhlmann_2010}, Stabl LASSO \cite{Hedou_Maric_2024}, RENT \cite{rent}, and HSIC LASSO \cite{Yamada_Tang_2018, Climente_2019}). Both univariable and multivariable filters were used, including information theory-based techniques (CIFE \cite{cife}, CMIM \cite{cmim}, DISR \cite{disr}, JMI \cite{jmi}) and similarity-based approaches (ReliefF \cite{relieff}, Fisher score \cite{data_perspective}), correlation-based clustering (hierarchical clustering \cite{hierarchical}), and statistical methods (Gini index, t-score, minimum redundancy maximum relevance (mRMR) \cite{data_perspective}, and adjusted p-values using Benjamini-Hochberg method \cite{Benjamini_Hochberg_1995}. Wrappers included two recursive elimination methods (LR-RFE, based on logistic regression coefficients \cite{Guyon_2002}, and RF-RFE, based on random forest importances \cite{Jiang_2004}), a forward selection method (forward RF \cite{Xia_Yang_2022}), and the SHAPley-based Shapicant \cite{shapicant}. Finally, random feature selection was included as a control. A complete description of all benchmarked methods is provided in Supplementary Table S1. 

Furthermore, before performing FS, \textit{ROOFS} allows users to conduct feature prefiltering using an iterative variance inflation factor (VIF) procedure. At each iteration, VIF of the $j_{\text{th}}$ feature is computed as:

$$\text{VIF}_j = \frac{1}{1 - R_j^2}$$

where $R_j^2$ stands for the coefficient of determination from regressing the $j_{\text{th}}$ feature onto all remaining predictors. The feature with the highest VIF is removed, and the VIF values are recomputed for the reduced predictor set. This process is repeated until no remaining feature exceeds a user-specified threshold. This allows to obtain a subset of predictors with reduced multicollinearity.

\subsubsection{Classifiers}

\textit{ROOFS} evaluates the predictive performance of signatures resulting from the FS step by training downstream classifiers on the corresponding post-selection datasets, allowing the use of any scikit-learn-compatible classifier. Since predictive performances can depend on the interactions between the FS method and the classifier, it is recommended to test a wide range of modeling approaches, such as linear models, tree-based methods, and ensembles. The classifiers used in the benchmark on PIONeeR data are listed in Supplementary Table S2.

\subsubsection{Optimism correction}

\begin{figure}[H]
    \centering
    \begin{subfigure}{1.05\textwidth}
        \centering
        \includegraphics[width=\textwidth]{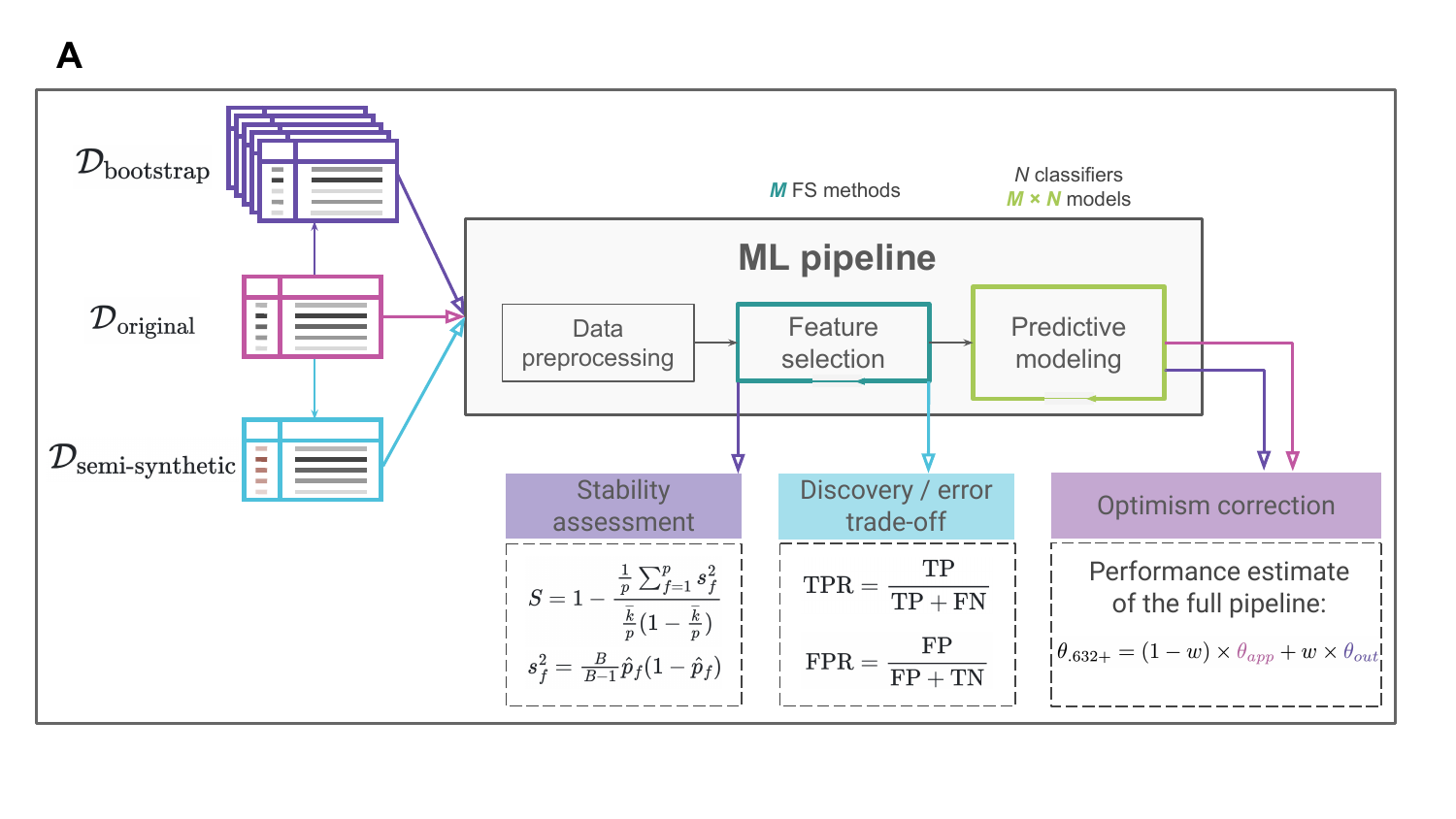}
    \end{subfigure}
    
    \begin{subfigure}{1.1\textwidth}
        \centering
        \includegraphics[width=\textwidth]{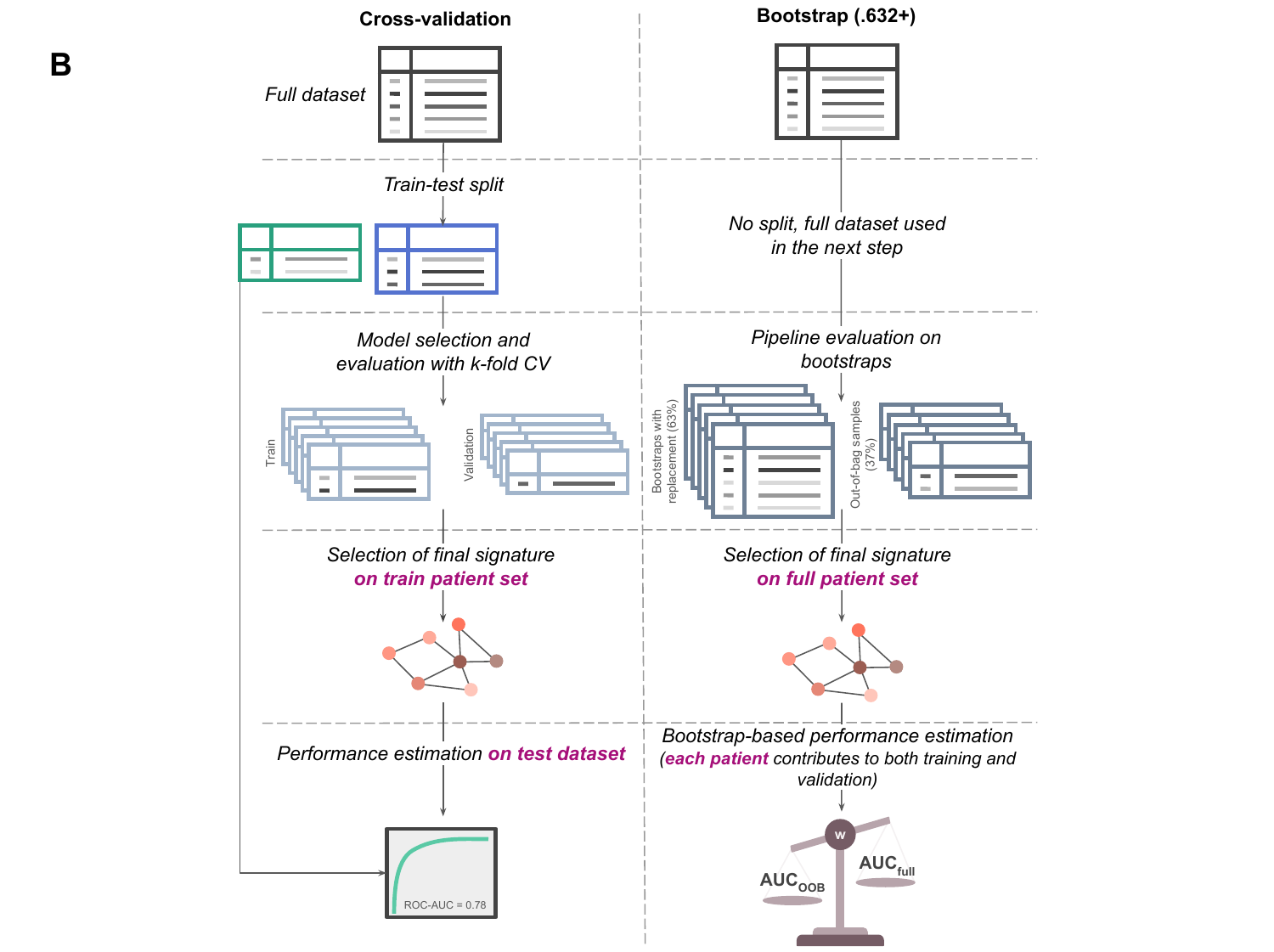}
    \end{subfigure}
    \vspace{0.05cm}
    \caption{A: Overview of the \textit{ROOFS} pipeline. \textit{ROOFS} generates $B$ bootstraps from the original dataset and feeds them, along with the original data, through an ML pipeline (imputation, normalization, feature selection, modeling) across all $M \times N$ method-classifier combinations. Additionally, \textit{ROOFS} supports generating a semi-synthetic dataset with a known ground-truth outcome, which can be fed through the same ML pipeline. This enables three parallel evaluations: stability (Nogueira's measure $S$ across bootstrap subsets), optimism-corrected predictive performance, and optionally discovery/error trade-off (TPR and FPR on semi-synthetic data). B: Comparison of a commonly used train/test split approach and .632+ optimism correction. The train/test split approach leaves a part of the data unused. \textit{ROOFS} recommends the .632+ bootstrap framework, which uses all patients for both signature derivation and validation: the full pipeline runs on $B$ bootstraps ($\sim 63\%$ of patients each), with out-of-bag samples ($\sim 37\%$) used for performance estimation. The final estimate $\text{AUC}_{\text{.632+}}$ is a weighted sum of apparent performance ($\text{AUC}_{\text{full}}$) and out-of-bag performance ($\text{AUC}_{\text{OOB}}$), with weight reflecting overfitting (see Methods).}
    \label{fig:scheme}
\end{figure}

In the commonly used train/test split framework, a portion of the data is held out for internal validation (Fig. \ref{fig:scheme}B). Model selection is performed on the training subset using cross-validation, and the final signature reported as the study outcome is derived from the training data. This practice arguably makes inefficient use of the limited data typically available in many studies. As pointed out by \citet{Riley_2020}, ``It is better to use all available data for model development (ie, avoid data splitting), with resampling methods (such as bootstrapping) used for internal validation."

Following this recommendation, \textit{ROOFS} evaluates predictive performance via optimism correction (OC), with every patient being used for both training and validation through bootstrapping with replacement or subsampling without replacement.

\newpage
\thispagestyle{empty} 
\newpage

Although we recommend using bootstrapping or subsampling, \textit{ROOFS} also supports FS benchmarking through train-test split and a single 5-fold cross-validation performed on the training set. However, this approach does not allow for the analysis of FS stability. 

Three OC methods are implemented: Harrell \cite{Harrell_Lee_Mark_1996}, .632 \cite{Efron_1983}, and .632+ \cite{Efron_Tibshirani_1997}. The following set of metrics are calculated: area under the receiver-operator curve (AUC), accuracy, sensitivity, specificity, positive predictive value, negative predictive value.

In Harrell's classical method, optimism is defined as the average difference between model performance on bootstrap samples (in training) and on the original dataset (in testing):

\[
O = \frac{1}{B} \sum_{b=1}^{B} \left( \theta_{\text{boot}}^{(b)} - \theta_{\text{orig}}^{(b)} \right),
\]

where \(\theta_{\text{boot}}^{(b)}\) is the training performance of the \(b\)-th bootstrap model, \(\theta_{\text{orig}}^{(b)}\) is the test performance of the same model on the original dataset, and \(B\) is the number of bootstrap replicates.

The final optimism-corrected performance is obtained by subtracting the average optimism from the apparent performance (i.e., the training performance on the full dataset):

\[
\theta_{\text{Harrell}} = \theta_{\text{app}} - O,
\]

where \(\theta_{\text{app}}\) denotes the apparent performance.

The .632 method accounts for the overlap between the bootstrap samples and the original data and computes $\theta_{\text{orig}}$ only on out-of-bag (OOB) samples (not included into the bootstrap, therefore not seen during training). The performance estimate is computed as
$$
\theta_{.632} = 0.368 \cdot \theta_{\text{app}} + 0.632 \cdot \theta_{\text{out}} ,
$$
where $\theta_{\text{out}} = \frac{1}{B} \sum_{b=1}^{B} \theta_{\text{OOB}}$. The weight $0.632$ corresponds to the approximate proportion of samples included in a bootstrap \cite{Chernick_LaBudde_2011}.

The .632+ method improves the .632 rule by modifying the weights in order to take into account the amount of overfitting:

$$
\theta_{\text{.632+}} = (1 - w) \cdot \theta_{\text{app}} + w \cdot \theta_{\text{out}},
$$

where 
$$
w = \frac{0.632}{1 - 0.368 \cdot R}
$$

and $R$ is the overfitting ratio computed as:

$$
R = \frac{\theta_{\text{app}} - \theta_{\text{out}}}{\theta_{\text{app}} - \gamma} ,
$$

where $\gamma$ corresponds to the predictive performance on the original dataset when the outcome is randomly permuted. When there is no overfitting (i.e., the apparent score is the same as the out-of-bag score), $R$ equals to 0, $w$ equals to 0.632, and the .632+ rule replicates the .632 rule. In the presence of overfitting (apparent score is greater than the out-of-bag score), $R \in (0, 1]$ and $w \in (0.632, 1]$. In a setting of complete overfitting, $w$ equals to 1 and $\theta_{\text{.632+}} = \theta_{\text{out}}$. The results reported in this paper were obtained using .632+ method as it has been demonstrated to have the lowest bias in multiple settings \cite{Iba_Shinozaki_Maruo_Noma_2021}. 

\subsubsection{FS stability}

The FS stability in \textit{ROOFS} is calculated from the bootstrap-derived feature sets using Nogueira's frequency-based measure (see \cite{Nogueira_2018} for details and a review of alternative stability metrics):
$$ S = 1 - \frac{\frac{1}{p}\sum_{f=1}^{p}s_f^{2}}{\frac{\overline{k}}{p}(1 - \frac{\overline{k}}{p})} $$

where $s_f^{2} = \frac{B}{B - 1}\hat{p}_f(1 - \hat{p}_f)$ is the unbiased sample variance of the frequency of selection $\hat{p}_f$ of $f^{th}$ feature over $B$ bootstraps and $\overline{k}$ is an average size of the selected subsets. $S$ takes values from $-\frac{1}{B - 1}$ to 1, with the lower bound asymptotically approaching 0 as $B$ approaches infinity. $S = 1$ indicates perfect stability when the same features are selected at each iteration. $S = 0$ corresponds to the case of random selection. Values $S < 0.4$ indicate poor stability, values $0.4 \leq S \leq 0.75$ are considered intermediate to good, and values $S > 0.75$ indicate excellent stability.

\subsubsection{Patient-level prediction stability}

Optionally, \textit{ROOFS} saves the decision thresholds and patient-level predicted probabilities from all built bootstrap models, allowing assessment of classification instability. For our benchmark on PIONeeR data, we computed a classification instability index (CII) for each individual as the proportion of bootstrap models that classified the individual differently from the apparent model \cite{Riley_Collins_2023}. The CII is expected to be highest for patients whose apparent predicted probability lies near the decision threshold, where small shifts in estimated risk change the classification, and near zero for individuals whose predictions are far from the threshold. We displayed this in a prediction instability plot, a scatter plot of the CII (y-axis) against each individual's apparent predicted probability (x-axis).

\subsubsection{Trade-off between true and false discoveries}

Complementary to the main benchmarking, \textit{ROOFS} evaluates each method’s ability to identify the ``true features" (i.e., the ones hypothesized as having generated the outcome) on a semi-synthetic dataset with a simulated outcome denoted $\tilde{y}$. With the true model being unknown, the real outcome $y$ cannot be used directly for this purpose, but the original $n \times p$ features matrix \textbf{X} can still be used. The synthetic outcome $\tilde{y}^{(i)}$ for observation $i$ is generated from a Bernoulli distribution with a parameter $p$ determined by a logistic function applied to a linear combination of
$S$ pre-selected features from the real dataset \textbf{X}:

$$
\tilde{y} \sim \text{Bernoulli}\!\left(p\right), \qquad
p = \frac{\exp\!\left(\sum_{j \in \mathcal{S}} \beta_j X_{j}\right)}{1 + \exp\!\left(\sum_{j \in \mathcal{S}} \beta_j X_{j}\right)},
$$

where $\mathcal{S}$ is the index set of the pre-selected true predictors and $\beta_j = 1 \ \forall \, j \in \mathcal{S}$.

The comparison of FS methods on a semi-synthetic data $\mathcal{D}_{\text{semi-synthetic}} = \{(x^i, \tilde{y}^i)\}_{i=1}^{n}, \quad x^i \in X, \, \tilde{y}^i\in \tilde{Y}$ allows assessing the trade-off between true and false discoveries, since the underlying true model is known. The true positive rate (TPR) – the proportion of true predictors correctly identified – represents the discovery power of an FS algorithm. The false positive rate (FPR) – the proportion of features not included in the true model but selected by an algorithm – shows the cost of discovery in terms of error. The variability in FPR and TPR is assessed by repeating the experiment, with the number of replicates specified by the user (set to 50 in our benchmark on PIONeeR data).

The function used to generate outcomes is simpler than real-world data-generating processes, creating a favorable scenario for FS and yielding optimistic TPR/FPR estimates. The package supports feature interactions and quadratic/cubic terms for testing more complicated settings.

In the experiment on the semi-synthetic dataset, fixed subset size methods were configured to select $k$ features matching the number of predictors in the true model. This gave them an advantage over other FS methods with respect to both true and false positives. Their TPR was lower bounded, as by design the selection conducted by these methods was not too stringent, and only a random or very inaccurate method would have obtained a low TPR. In turn, their FPR was upper bounded, as these methods could not select more than $k$ features.

\subsection{Additional functionalities}
\label{subsec:highlighted-functionalities}

\begin{enumerate}

\item Benchmarking a handcrafted FS method

\vspace{0.15cm}

A key feature of the package is that new methods can be easily incorporated into the benchmark. Users developing custom FS methods can therefore evaluate them alongside those already implemented in \textit{ROOFS}. This is possible by enveloping new methods into Python functions and passing them as arguments to the \texttt{fs\_builder} method of the \texttt{Optimism} class (see an example in the Jupyter notebook provided in the repository). Similarly, existing methods that are not included in the package can be evaluated in this way.

\vspace{0.15cm}

\item Control FS methods

\vspace{0.15cm}

It is recommended to compare the benchmarked FS algorithms with the control methods provided: random FS (lower performance bound), full model without any FS (upper bound, as all information is preserved), and optionally predefined fixed signatures.

\vspace{0.15cm}

\item Choosing an optimal subset size

\vspace{0.15cm}

As described above, some of the FS algorithms in \textit{ROOFS} select a fixed number of features and therefore require a parameter that specifies the desired subset cardinality. Several approaches can be used to choose an appropriate value of this parameter. First, \textit{ROOFS} provides an automated procedure to guide the choice of subset size by jointly optimizing predictive performance, stability, and sparsity (see Supplementary information, subsection "Choosing an optimal subset size"). Second, a useful baseline can be derived from the methods that determine subset size automatically. Finally, a 10-EPV (events per variable) rule can be applied, according to which each selected feature requires at least ten observations in the minority outcome class \cite{Peduzzi_Concato_Feinstein_Holford_1995}.

\vspace{0.15cm}

\item Robustness of individual features

\vspace{0.15cm}

In addition to overall stability assessed via Nogueira's measure, we recommend examining per-feature selection frequencies across bootstrap resamples. For each feature, \textit{ROOFS} computes its selection frequency within each FS method. It serves as an indicator of the signal strength: features that are weakly associated with the outcome appear in only a small fraction of bootstrap models, whereas features with a strong signal are consistently selected in a large proportion of bootstrap models and across multiple FS methods. This enables distinguishing between features that are stable and predictive and those that may be artifacts of a specific algorithm or over- or undersampling of particular patient subclusters. Per-feature selection frequencies also shed light on the variables outside the final signature that may carry predictive information, as indicated by high selection frequencies across bootstrap samples, and which may, therefore, require further investigation.

\vspace{0.15cm}

\item Method consensus

\vspace{0.15cm}

Furthermore, for each feature, \textit{ROOFS} reports the number of methods that select it with high frequency (where the frequency threshold is user-defined; set to 50\% in our experiments). Features selected consistently by multiple methods, especially by methods from different families, can be interpreted as more reliable.

\vspace{0.15cm}





\end{enumerate}

\section{Results} 

\subsection{PIONeeR dataset}

The development of \textit{ROOFS} was motivated by the following real-world problem: identifying robust biomarkers of resistance to immunotherapy in lung cancer. To this end, we relied on the PIONeeR clinical trial dataset, derived from a prospective, multicenter cohort of patients (17 centers in France) with advanced or recurrent NSCLC \cite{Barlesi_2026}. The study is registered at ClinicalTrials.gov (NCT03493581), registration date is 2019-02-05.

Patients received either frontline combination therapy with platinum-based chemotherapy plus anti-PD-(L)1 immune checkpoint inhibitors (ICIs) or second-/third-line monotherapy with anti-PD-(L)1 ICIs following progression on prior platinum-based chemotherapy. The dataset integrated multi-modal data from $n = 435$ patients, comprising a total of $p = 374$ variables. It included clinical and demographic variables (e.g., sex, age, $p = 10$), routine blood tests (e.g., blood counts, biochemistry, $p = 49$), tumor multiplex immunohistochemistry markers (e.g., immune cells, $p = 159$), and circulating immune and vasculophenotyping markers, which included both flow cytometry data ($p = 141$) and soluble ($p = 15$) markers. The dataset contained 30.8\% of missing values, which were imputed using the median or mode. For methods unable to handle missing data, imputation was performed on each bootstrap dataset before FS; for methods that handle missing values directly, it was performed before model training. The outcome to predict was primary resistance to immunotherapy, defined as disease progression within 6 months of treatment \cite{SITC_2023}.

\subsection{A LASSO-derived signature shows modest stability}

\begin{figure}[ht]
    \centering
    \includegraphics[width=\textwidth]{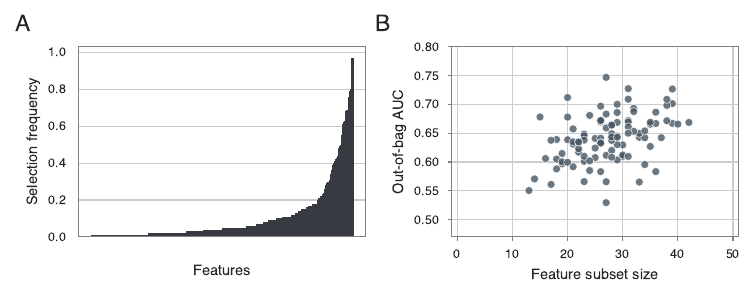}
    \caption{Instability of LASSO on the full PIONeeR dataset ($p = 374$). \textbf{A:} Bootstrap selection frequencies for features selected by LASSO in at least 1 bootstrap sample. \textbf{B:} Variability in selected subset sizes and out-of-bag AUC.}
    \label{fig1}
\end{figure}

We first applied our benchmarking pipeline, as described in Methods and illustrated in Fig. \ref{fig:scheme}A, using classical LASSO (Least Absolute Shrinkage and Selection Operator) as a baseline FS method. LASSO remains a standard FS approach in biomedical data analysis due to its simplicity of interpretation and computational efficiency, but it has been shown to be inconsistent (i.e., unable to recover a true model) in the presence of linear dependencies among features, which is common in high-dimensional problems \cite{Zhao_Yu_2006}. On the PIONeeR data, LASSO showed poor stability (S = 0.34, Fig. \ref{fig1}A) and considerable variability in features selected across bootstrap datasets, with 255 out of 374 features selected in at least 1 bootstrap. LASSO-selected feature sets varied substantially in size, ranging from 13 to 42 predictors, and the corresponding post-selection models, trained using classical ML classifiers, achieved out-of-bag AUC values ($\theta_{\text{out}}$) between 0.49 and 0.78 (Fig. \ref{fig1}B). The apparent LASSO model selected 25 features, of which only 12 were selected with high confidence (defined here as being chosen in at least 50\% of bootstrap models), and none were selected across all bootstraps. Low selection frequency of the remaining 13 features raised concerns about their potential clinical relevance and the generalizability of the signature.

\subsection{Iterative VIF increases FS stability}

\begin{figure}[ht]
    \centering
    \includegraphics[width=0.95\textwidth]{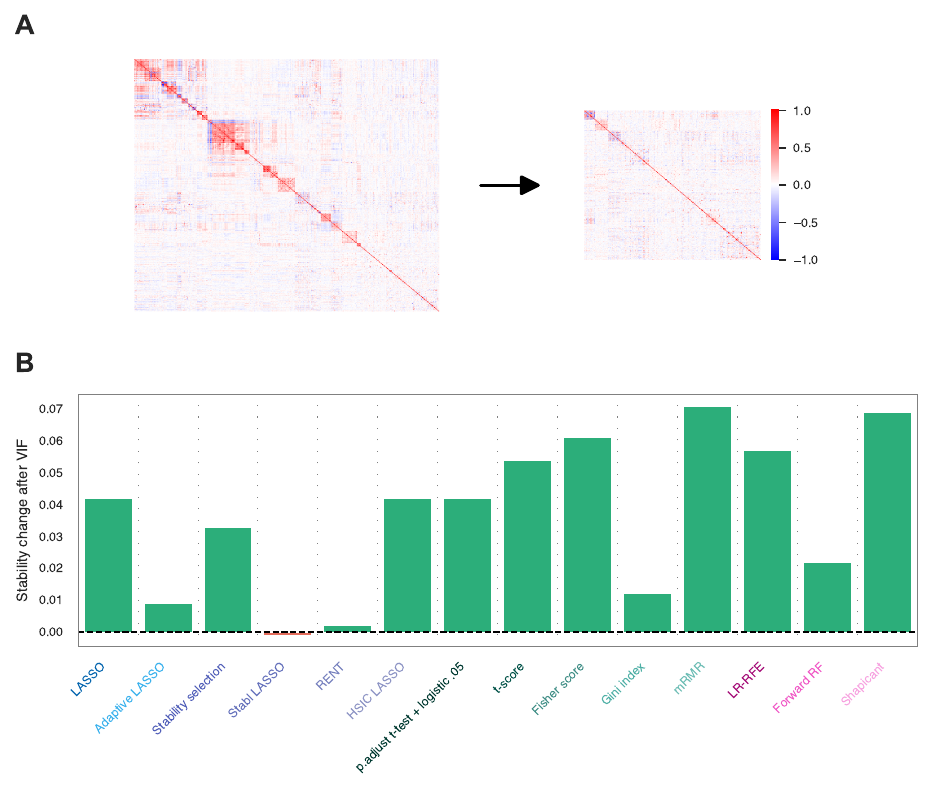}
    \caption{Improvement in FS stability following multicollinearity reduction with VIF pre-filtering compared to the original dataset. \textbf{A:} Correlation structure of the dataset before and after applying VIF pre-filtering. \textbf{B:} Absolute change in FS stability across representative FS methods.}
    \label{fig2}
\end{figure}

Since LASSO instability could be partially attributed to multicollinearity among predictors, we next applied iterative VIF pre-filtering to the full data (374 features, imputed) to derive a feature set with lower redundancy and evaluate its impact on FS stability. This produced a reduced dataset of 214 features (Fig. \ref{fig2}A). The mean $R^2$ obtained from a regression of each feature onto all other predictors decreased from 0.91 (range: 0.41-0.99) in the full dataset to 0.63 (range: 0.22-0.79) following VIF pre-filtering.

To evaluate whether this reduction improved FS stability, we conducted two separate benchmarks of 23 FS methods on the full and reduced datasets. VIF pre-filtering led to a small increase in stability of the majority of FS methods (Fig. \ref{fig2}B), likely driven by the fact that fewer features were used interchangeably across bootstrap models. Additionally, the majority of FS methods showed no loss in AUC, indicating that VIF pre-filtering did not discard features carrying indispensable predictive information. Based on this improvement, all subsequent results are reported on the post-VIF dataset.

\subsection{Statistical filters offer the best predictive performance and selection stability}

\begin{figure}[ht]
    \centering

    \includegraphics[width=0.9\textwidth]{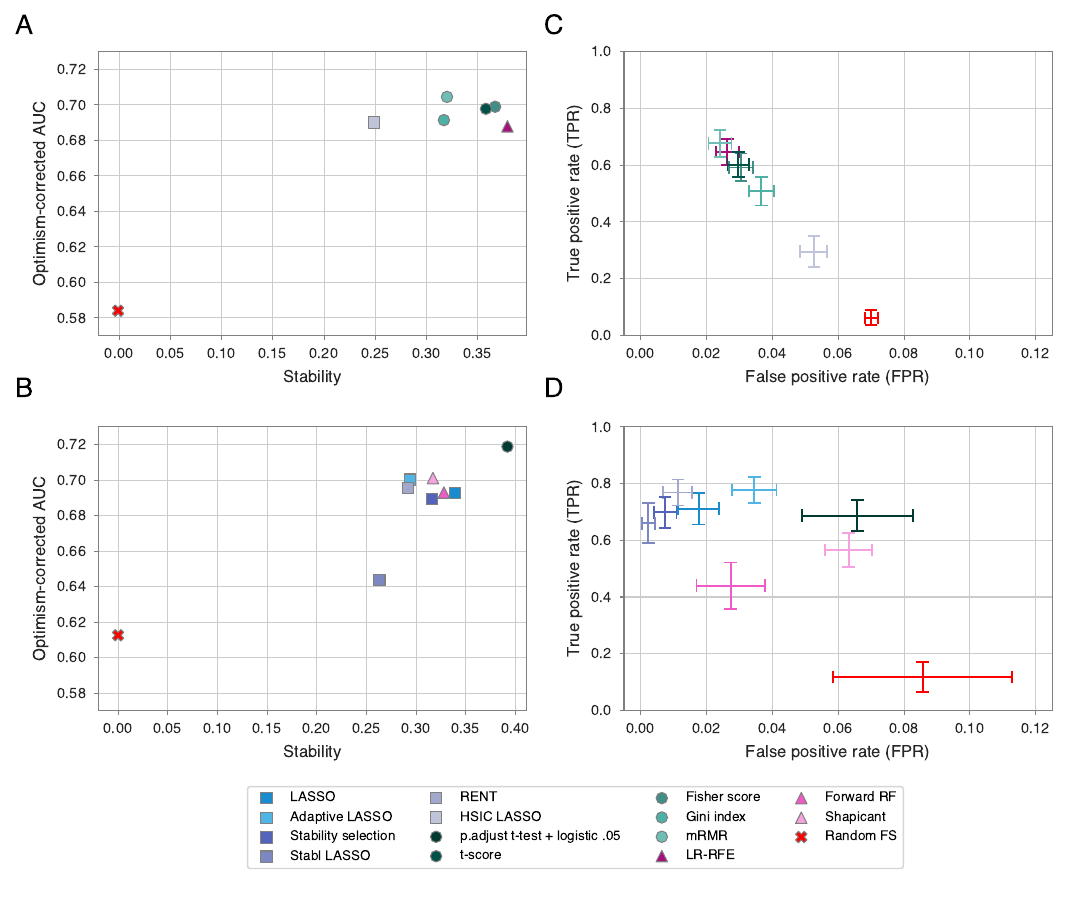}
    \caption{Performance of representative FS methods from different algorithmic families (top: FS methods with fixed, user-defined subset size; bottom: FS methods with varying subset size dependent on algorithm thresholds). Blue: embedded methods; violet-blue: resampling-based ensemble methods; green: filters; rose: wrappers. \textbf{A, B:} Stability and AUC in the experiment on the post-VIF PIONeeR dataset. \textbf{C, D:} Trade-off between discovery and error in the experiment on the semi-synthetic PIONeeR dataset with a simulated outcome. The true positive rate (TPR) shows the proportion of true predictors correctly identified by each FS method, while the false positive rate (FPR) shows the proportion of non-informative features incorrectly selected. Error bars represent one-half of the standard deviation (SD).}
    \label{fig3}
\end{figure}

On the reduced post-VIF dataset, filters were generally the most stable and predictive family of FS methods. This behavior is expected, because filter methods are deterministic and score features using a single criterion. Unlike more complicated algorithms involving stochastic components or performance-based optimization, they introduce no additional source of variability beyond that arising from bootstrap resampling. Nevertheless, some variability was observed within the group (Fig. \ref{fig3}A-B; Supplementary fig. S2A-B). Statistical filters achieved optimism-corrected AUC values ranging from 0.69 to 0.72, and several of them demonstrated improved stability as compared to LASSO, for example, t-score (S = 0.36) and p.adjust (S = 0.39) (Table \ref{tab:results}). In contrast, JMI and DISR, methods based on information theory scores, achieved top stability values (S = 0.77 and 0.72, respectively) but lower predictive performance (optimism-corrected AUC = 0.58). DISR, CIFE, CMIM and JMI frequently omitted known biomarkers (such as LDH or C-reactive protein), which explains their poor predictive performance. Notably, both JMI (S = 0.77) and CMIM (S = 0.33) rely on the same ranking criterion and differ only in that JMI divides the sum of feature redundancy terms by the cardinality of the selected subset, which places greater weight on the mutual information between the feature and the outcome. This suggests that JMI stability may stem from the high stability of the outcome-feature mutual information term. Its poor predictive performance, however, indicates that this criterion, despite being stable, is not well suited to the PIONeeR data. Additionally, information theory methods require discretization of continuous variables, which might have led to the loss of information.

Embedded methods exhibited the lowest stability among other groups at comparable predictive performance. Surprisingly, with the exception of RENT, embedded resampling-based ensembles designed to improve the consistency of FS (Bolasso, stability selection, Stabl LASSO, and HSIC LASSO) outperformed standard LASSO in neither stability nor AUC. Their stringent selection (with the exception of HSIC LASSO) may account for their lower predictive performance, but it does not explain the lack of improvement in stability (S = 0.34, 0.32, 0.26, and 0.25, respectively, vs LASSO S = 0.34). The instability of these methods could be attributed to the variability behind the resampling procedures on which they rely, although this requires further investigation. The performance of the wrapper methods was intermediate, with AUC (0.69-0.71) and stability (0.32-0.39) falling between filter methods (the best) and embedded methods (the worst).

\begin{table}[ht]
\setlength{\tabcolsep}{3pt}
\caption{Results of FS benchmarking on the post-VIF PIONeeR dataset ($n = 435, p = 214$)}
\centering
\begin{tabular}{llllll}
\toprule
                         Method &  AUC &  Stability &   FPR &   TPR & Run time (s)\\
\midrule
                          LASSO & 0.69 \textsuperscript{St} &       0.34 & 0.02 ± 0.01 & 0.71 ± 0.11 & 0.27 ± 0.02\\
                 Adaptive LASSO & \textcolor{Cyan4}{0.7} \textsuperscript{St} &       0.3 & 0.03 ± 0.01 &  \textcolor{Cyan4}{\textbf{0.78 ± 0.09}} & \textcolor{Cyan4}{0.24 ± 0.04} \\
                Exclusive LASSO & 0.67 \textsuperscript{St} &       \textcolor{Cyan4}{0.37} & 0.17 ± 0.02 & \textcolor{Cyan4}{0.75 ± 0.11} & 0.59 ± 0.06 \\
                \midrule
                    Bolasso & 0.69 \textsuperscript{GB} &       0.34 & \textcolor{Cyan4}{0.01 ± 0.01} & 0.68 ± 0.14 & 17.48 ± 0.71\\
              Stability selection & 0.69 \textsuperscript{SVC} &       0.32 & \textcolor{Cyan4}{0.01 ± 0.01} & 0.7 ± 0.11 & 8.10 ± 0.6 \\
                    Stabl LASSO & 0.64 \textsuperscript{St} &       0.26 & \textbf{\textcolor{Cyan4}{0.00 ± 0.00}} & 0.66 ± 0.14 & 47.36 ± 5.2 \\
                           RENT & \textcolor{Cyan4}{0.7} \textsuperscript{St} &       0.29 &  \textcolor{Cyan4}{0.01 ± 0.01} & \textcolor{Cyan4}{0.77 ± 0.09} & 46.48 ± 4.23 \\
                        HSIC LASSO & 0.69 \textsuperscript{Ex} &       0.25 & 0.05 ± 0.01 & 0.3 ± 0.11 & 0.69 ± 0.06 \\
\midrule
                           CIFE & 0.61 \textsuperscript{SVC} &       \textcolor{Cyan4}{0.42} & 0.06 ± 0.00 & 0.15 ± 0.06 & 3.4 ± 0.14 \\
                           CMIM & 0.61 \textsuperscript{LR} &       0.33 & 0.05 ± 0.01 & 0.36 ± 0.1 & 3.62 ± 0.15\\
                           DISR & 0.58 \textsuperscript{MLP} &       \textcolor{Cyan4}{0.72} &  0.04 ± 0.01 &  0.41 ± 0.1 & 10.11 ± 0.23 \\
                            JMI & 0.58 \textsuperscript{LDA} &       \textbf{\textcolor{Cyan4}{0.77}} &  0.05 ± 0.01 &  0.37 ± 0.1 & 3.79 ± 0.07 \\
                        ReliefF & 0.68 \textsuperscript{LR} &       \textcolor{Cyan4}{0.49} & 0.06 ± 0.01 & 0.22 ± 0.08 & \textcolor{Cyan4}{0.26 ± 0.03} \\
                    Fisher score & \textcolor{Cyan4}{0.7} \textsuperscript{LDA} &       \textcolor{Cyan4}{0.37} & 0.03 ± 0.01 & 0.59 ± 0.1 & \textcolor{Cyan4}{0.19 ± 0.03} \\
                     Gini index & 0.69 \textsuperscript{Ex} &       0.32 & 0.04 ± 0.01 & 0.51 ± 0.11 & 1.00 ± 0.04 \\
                        t-score & \textcolor{Cyan4}{0.7} \textsuperscript{GB} &       \textcolor{Cyan4}{0.36} & 0.03 ± 0.01 & 0.60 ± 0.09 & \textcolor{Cyan4}{\textbf{0.11 ± 0.03}} \\
                           mRMR & \textcolor{Cyan4}{0.7} \textsuperscript{LDA} &       0.32 & 0.02 ± 0.01 & 0.68 ± 0.1 & 0.82 ± 0.03 \\
        Hierarchical clustering & \textcolor{Cyan4}{0.7} \textsuperscript{GB} &       0.23 &  0.04 ± 0.01 &  0.47 ± 0.09 & 9.44 ± 0.47 \\
 p.adjust t-test + logistic .05 & \textbf{\textcolor{Cyan4}{0.72}} \textsuperscript{GB} &       \textcolor{Cyan4}{0.39} & 0.07 ± 0.03 & 0.69 ± 0.11 & 1.95 ± 0.09 \\
 \midrule
                         LR-RFE & 0.69 \textsuperscript{LR} &      \textcolor{Cyan4}{0.38} &  0.03 ± 0.01 &  0.65 ± 0.09 & \textcolor{Cyan4}{\textbf{0.11 ± 0.03}} \\
                         RF-RFE & \textcolor{Cyan4}{0.71} \textsuperscript{GB} &       \textcolor{Cyan4}{0.39} & 0.03 ± 0.02 & 0.48 ± 0.12 & 52.27 ± 0.30 \\
                     Forward RF & 0.69 \textsuperscript{LDA} &       0.33 & 0.03 ± 0.02 & 0.44 ± 0.16 & 16.29 ± 0.13 \\
                      Shapicant & \textcolor{Cyan4}{0.7} \textsuperscript{GB} &       0.32 & 0.06 ± 0.01 & 0.57 ± 0.12 & 25.96 ± 0.20 \\
\midrule
                      Random FS (fixed size) & 0.58 \textsuperscript{HistGB} &      0 &             0.07 ± 0.00 & 0.06 ± 0.05 &\\
                      Random FS & 0.61 \textsuperscript{RF} &      0 &             0.09 ± 0.05 & 0.12 ± 0.1 &\\
                      Full data & 0.71 \textsuperscript{GB} &       1 &             & & \\
\bottomrule
\end{tabular}
\label{tab:results}
\footnotesize
\parbox{0.95\textwidth}{%
 AUC is reported for the optimal FS method-classifier pair across all tested combinations. FPR and TPR were evaluated over 50 repetitions. Values are presented as point estimates or mean ± SD (standard deviation); run times correspond to the average duration of the FS step (excluding predictive modeling) across bootstraps. \textcolor{Cyan4}{Cyan indicates better performance than that of LASSO.} 
 GB: Gradient boosting classifier; HistGB: Histogram-based gradient boosting classifier; St: stacking classifier; Ex: extra trees classifier; Bag: bagging classifier; LR: logistic regression; SVC: support vector classifier; LDA: linear discriminant analysis; MLP: multi-layer perceptron; LASSO: least absolute shrinkage and selection operator; HSIC: Hilbert-Schmidt independence criterion; RENT: repeated elastic net technique; CIFE: conditional infomax feature extraction; CMIM: conditional mutual information maximization; DISR: double input symmetrical relevance; JMI: joint mutual information; mRMR: minimum Redundancy Maximum Relevance; LR-RFE: logistic regression-based recursive feature elimination.
}
\end{table}

Among all methods, the p.adjust, a statistical filter that combined t-test and logistic regression adjusted p-values at a 0.05 threshold (22-feature signature on the full data), achieved the best balance between stability (S = 0.39) and performance (AUC = 0.72) (Fig. \ref{fig3}A). The gradient boosting (GB) model trained following the p.adjust FS step achieved predictive performance equivalent to that of the GB model trained on the full feature set (214 features) without FS (AUC = 0.71). Note that the strong performance of the FS-free GB model is likely attributed to the restriction of the depth of tree-based classifiers in the benchmark to 2, which added an additional internal layer of FS.

Overall, most benchmarked methods achieved optimism-corrected AUC values comparable to LASSO, with statistical filters and wrappers showing modest improvements and information-theory filters performing worse. However, several methods with stronger predictive performance also demonstrated higher stability than LASSO, supporting the idea that it is possible to select features more consistently while maintaining similar predictive performance \cite{Nogueira_2018}. Across all methods, stability and AUC showed a weak negative association (Spearman's $r = -0.304$, $p = 0.158$), explained by the poor performance of stable information-theoretic methods. Excluding CIFE, CMIM, DISR, and JMI shifted the association to zero ($r = 0.06$).

\subsection{Predictive performance is driven by the recovery of true predictors rather than by stringent control of false discoveries}

The second part of the benchmark was conducted on a semi-synthetic dataset to assess the ability of FS methods to discover informative predictors when a true model is known. As detailed in Implementation, a synthetic outcome variable was simulated with a linear model from 15 randomly chosen features from the real PIONeeR dataset, and FS methods were applied to the original dataset combined with this simulated outcome. Performance was assessed by measuring the true positive rate (TPR) and false positive rate (FPR) (Fig. \ref{fig3}C-D, Supplementary fig. S2C-D).

Embedded methods achieved the best balance between these two metrics (Table \ref{tab:results}). Ensemble approaches that control for false discoveries (Bolasso, stability selection, Stabl LASSO, and RENT) showed the lowest FPR, as expected, while avoiding excessive selection stringency and maintaining relatively high TPR. RENT, based on repeated elastic net fits, was the best method overall, with TPR = 0.77 ± 0.09 and FPR = 0.01 ± 0.01.

By construction, fixed-size methods exhibited an inverse relationship between TPR and FPR, as true discoveries constrained the number of false positives (Table \ref{tab:results}, Fig. \ref{fig3}A). Among them, mRMR and LR-RFE demonstrated the best balance between discovery and error (TPR = 0.68 ± 0.1 and 0.60 ± 0.09, FPR = 0.02 ± 0.01 and 0.03 ± 0.01, respectively). Information-based methods and reliefF showed lower discovery ability (TPR range: 0.15-0.41).

Overall, a low FPR did not necessarily translate into better predictive performance, while higher TPR values were associated with a higher AUC (Spearman's $r = 0.44$, $p = 0.036$). This aligns with the intuitive notion that strong generalization depends on identifying a sufficiently rich set of informative predictors, rather than solely minimizing false positives.

Linear FS methods achieved the strongest overall performance, which is consistent with the linear data-generating function used in the simulation. However, additional experiments demonstrated that method rankings varied when the same outcome-generating function was applied to different datasets, confirming the relevance of dataset-specific benchmarking. The results in different scenarios we tested showed that adaptive LASSO maintains an excellent balance between discovery and sparsity and mRMR demonstrates moderate but robust performance across all settings, including non-linear data-generating functions. p.adjust consistently recovers a high proportion of true predictors, often at the cost of high false discoveries, achieving low FPR only in the absence of feature redundancy, consistent with evidence that correlations inflate false discoveries in multiple hypothesis testing settings \citet{Kanduri_2025}.

Table \ref{tab:results} additionally compares computational efficiency across FS methods. Most filter and embedded methods required $\leq10$ seconds per bootstrap iteration (and $\leq1$ second for some). Wrapper methods and resampling-based approaches were substantially more demanding, with computation time approaching 1 minute per bootstrap iteration.

\subsection{p.adjust filter achieves a more stable and predictive signature}

As described above, p.adjust, a filter method based on a union of features selected by a t-test and logistic regression controlled for PD-L1 at a 0.05 Benjamini-Hochberg-adjusted threshold, achieved the best predictive performance (optimism-corrected AUC of 0.72) and the best selection stability (S = 0.39) among the well-performing methods. 

\begin{figure}[ht]
    \centering
    \includegraphics[width=0.9\textwidth]{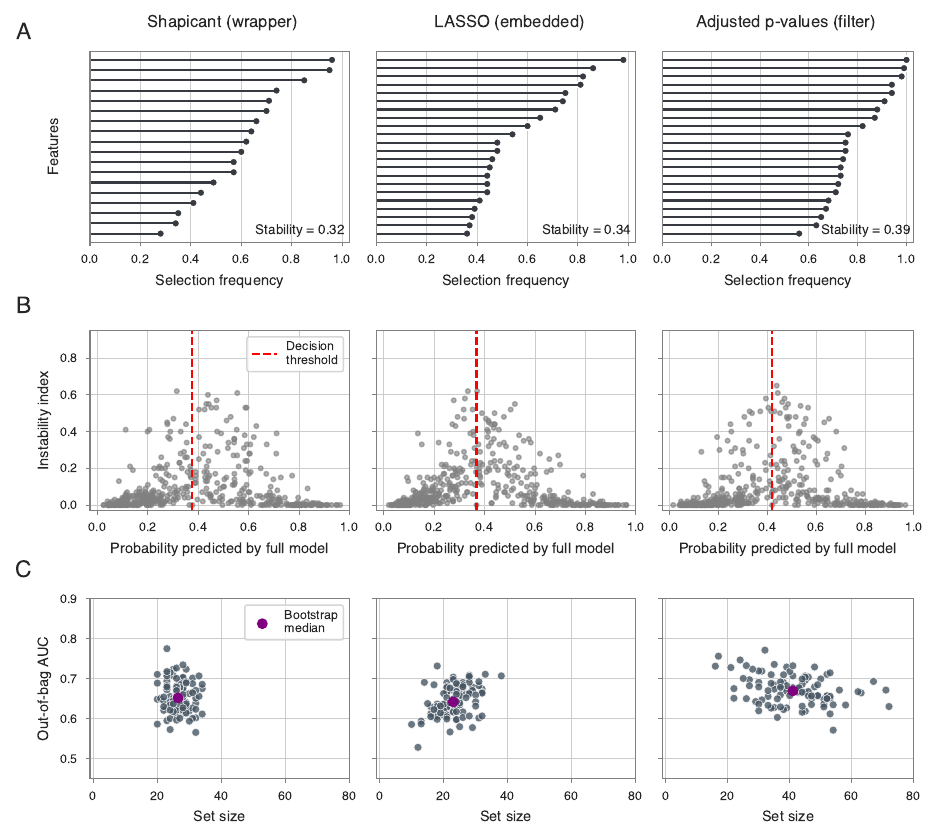}
    \caption{Comparison of performance between Shapicant (representative wrapper method), LASSO (baseline embedded method), and p.adjust (the FS approach selected using \textit{ROOFS}). \textbf{A:} Selection frequency of signature features. \textbf{B:} Instability index, defined as the proportion of bootstrap models leading to a different classification decision for a given patient compared to the full model \cite{Riley_Collins_2023}. \textbf{C:} Out-of-bag AUC ($\theta_{\text{OOB}}$; see Methods) as a function of selected subset size across bootstrap models.}
    \label{fig4}
\end{figure}

In addition to global stability, we assessed the robustness of individual signature features. Unlike standard LASSO, a baseline embedded FS method, and Shapicant, a representative of wrapper methods, all p.adjust-derived features were selected in $>50\%$ of bootstrap samples, increasing the reliability of the resulting signature (Fig. \ref{fig4}A).

We also evaluated the stability of the p.adjust models at the individual patient level as measured by the instability index \cite{Riley_Collins_2023} – the proportion of bootstrap models that classify a given patient differently compared to the full model (Fig. \ref{fig4}B). Ideally, the instability index should be elevated only for patients whose predicted probabilities lie close to the decision threshold of the full model. All models lacked this behavior. However, overall p.adjust improved patient-level stability: only 24.4\% of patients had an instability index exceeding 0.1, compared to 27.1\% for Shapicant and 31.7\% for LASSO.

Fig. \ref{fig4}C shows the distribution of bootstrap out-of-bag AUC values for the three methods. Although p.adjust achieved higher predictive performance, it selected substantially larger feature subsets, despite the signature obtained from the full dataset comprised only 22 features. This increase in subset size is attributed to the bootstrap resampling procedure, as it was no longer observed when the same pipeline was applied to datasets derived by subsampling to 80\% of the original sample size without replacement.

\section{Discussion}

This paper introduced \textit{ROOFS}, a Python framework designed to help researchers identify an FS method adapted to their predictive task. FS is a methodologically mature field, with extensive advances in algorithmic development \cite{data_perspective}, stability assessment \cite{Nogueira_2018}, and false discovery control \cite{Hedou_Maric_2024}. However, many of the theoretical developments have not been fully adopted in applied biomedical research, where practitioners often rely on a relatively narrow set of familiar methods. Moreover, FS is known to be data-dependent, with method performance varying across datasets and no single approach being universally optimal \cite{James_2017, Somol_2010, Drotar_2015}. Nevertheless, FS methods are often applied with little consideration of their suitability for the specific dataset at hand. \textit{ROOFS} addresses these gaps by integrating the evaluation of over 30 FS methods from different algorithmic families into a single pipeline that requires low to zero additional programming. To facilitate method comparison, \textit{ROOFS} generates automated reports summarizing FS performance metrics and measures of individual feature robustness. A key element of \textit{ROOFS} is its emphasis on stability, a property of FS that is frequently ignored in biomedical studies and in the development of new FS approaches, despite its importance for reproducibility.

To our knowledge, no previously released package combines multiple pre-implemented FS methods with stability assessment and optimism correction framework for model evaluation. The closest existing tool is the R package \textit{pminternal} (\underline{https://github.com/stephenrho/pminternal}), which estimates optimism of the user-defined model functions, including those that contain an FS step. However, it does not provide any pre-implemented FS methods.

We demonstrated the utility of \textit{ROOFS} through application to the dataset from the PIONeeR clinical trial, focused on predicting resistance to anti-PD-(L)1 immunotherapy in advanced lung cancer. It enabled identification of the adjusted p-value filter, defined as the union of features selected by a t-test and by logistic regression contolled for PD-L1 at a 0.05 threshold, as the FS method that offered the best stability, confidence in selected features, and predictive performance, with acceptable TPR and FPR. The overall findings of this benchmark study align with previous benchmarks where simple statistical filters outperformed more complicated ML approaches \cite{Haury_Gestraud_Vert_2011}. Therefore, we recommend starting the analysis of new datasets by benchmarking filter methods, given their additional advantage of computational efficiency. \textit{ROOFS} was also recently successfully applied to the analysis of the SChISM clinical study (Size CfDNA Immunotherapy Signature Monitoring, NCT05083494). It enabled identification of a nine-variable signature derived from fragmentomic circulating-free DNA data with improved predictive performance for resistance to therapy (AUC = 0.89 ± 0.033, PPV = 0.69 ± 0.075) \cite{schism_preprint}.

This work has several limitations. First, \textit{ROOFS} currently supports only classification tasks; future work will extend the framework to regression and survival tasks. Second, a full benchmark can require considerable runtime. While all implemented methods are scalable to moderately high-dimensional data, analysis of data with thousands of features may benefit from reducing the computational burden by excluding wrapper and ensemble methods and by limiting the number of classifiers used in downstream prediction. 

A related limitation is that \textit{ROOFS} currently supports hyperparameter tuning only for subset size. Including multi-parameter grid search in a resampling-based benchmarking framework would substantially increase runtime and may make the procedure impractical. Therefore, all implemented FS methods are evaluated using fixed parameters, which are expected to achieve good performance. Users who would like to test other settings can do so by modifying the package source code.

Finally, in the benchmark on the PIONeeR data, the synthetic outcome used to evaluate TPR and FPR was generated using only a linear model. This simplification may lead to inaccurate estimates and may bias comparisons in favor of linear FS methods. That said, we found that a single data-generating model was adequate for obtaining an optimistic baseline estimate of TPR and FPR. Additional experiments confirmed the relevance of dataset-specific comparison by showing that applying the same synthetic outcome-generating function to different datasets produces different FS method rankings. Importantly, since \textit{ROOFS} supports inclusion of nonlinear terms into a model, this is a limitation of a given benchmark, not the package itself.

\section{Conclusion}

\textit{ROOFS} can serve as a practical tool for researchers working in settings where reproducibility is crucial. By enabling comprehensive benchmarking, \textit{ROOFS} and similar tools have the potential to improve the reliability of biomarker discoveries and increase the translational value of clinical predictive models.

\vspace{0.15cm}

\textbf{Acknowledgements}

\vspace{0.15cm}

The authors gratefully acknowledge the support of the APHM, which sponsored the PIONeeR clinical study. Its role was to control the appropriateness of ethical and legal considerations for all centers and to perform the monitoring of the consents signed and the clinical data recorded and coded as part of the study. The authors are grateful to all the patients and their families, as well as all the investigators, for their participation in the study.

The authors are grateful to Linh Nguyen Phuong, PhD, and Mohamed Boussena for helpful comments.

\section*{Declarations}

\textbf{Author contribution}

\vspace{0.15cm}

A.B. and S.B. drafted the manuscript. A.B., S.B. and P.D. contributed to conceptualization and performed data analysis. A.B., S.C., and S.B. developed the software. F.M., L.G., and F.B. contributed to data acquisition. F.B. and S.B. secured funding. S.B. supervised the work. All authors reviewed and approved the final manuscript.

\vspace{0.15cm}

\textbf{Funding}

\vspace{0.15cm}

This work was supported by the French National Research Agency (ANR) under the France 2030 investment plan through grants ANR-17-RHUS-0007 and ANR-22-PESN-0017, the latter awarded within the DIGPHAT project.

This work was supported by a partnership of Aix-Marseille Université (AMU), Assistance Publique Hôpitaux de Marseille (APHM), Centre National de La Recherche Scientifique (CNRS), Institut National de la Santé et de la Recherche Médicale (INSERM), Centre Léon Bérard (CLB), Institut Paoli Calmettes (IPC), Gustave Roussy (GR), AstraZeneca (AZ), Veracyte (VERA), Innate Pharma (IPH) \& ImCheck Therapeutics (ICT), and initiated by Marseille Immunopole. 

\vspace{0.15cm}

\textbf{Data availability}

\vspace{0.15cm}

The PIONeeR data underlying this article cannot be shared publicly due to patient privacy reasons. Two publicly available datasets are included in the package for use in tutorials and software testing. The SChISM dataset was obtained from Zenodo (version 1; record 17854973; DOI: 10.5281/zenodo.17854973; link: \url{https://zenodo.org/records/17854973}) \cite{schism_zenodo, schism_preprint}. The Immuno-Curie dataset was obtained from Zenodo (version 1.0.0; record 14293431; DOI: 10.5281/zenodo.14293431; link: \url{https://zenodo.org/records/14293431}) \cite{curie_dataset, curie_zenodo}. A synthetic dataset generated for this package is also included.

\vspace{0.15cm}

\textbf{Competing interests}

\vspace{0.15cm}

The authors declare no conflict of interest.

\vspace{0.15cm}

\textbf{Ethics approval}

\vspace{0.15cm}

The PIONeeR study was conducted in accordance with the Helsinki declaration, French laws and regulations and the International Conference on Harmonization (ICH) E6 Guideline for Good Clinical Practice. The study was approved by the French ethics committee (Comité de Protection des Personnes Ouest II Angers, no. 2018/08) and the French drug and device regulation agency (Agence Nationale de Sécurité du Médicament, no. 2018020500208). 

\vspace{0.15cm}

\textbf{Consent to participate}

\vspace{0.15cm}

Informed consent was obtained from each participant before any study procedure. 

\vspace{0.15cm}

\textbf{Consent to publish}

\vspace{0.15cm}

Not applicable.

\section*{Availability and Requirements}

Project name: ROOFS (RObust biOmarker Feature Selection)

Project home page: \url{https://gitlab.inria.fr/compo/roofs}

Operating system(s): Platform-independent

Programming language: Python

Other requirements: Python 3.10; optional (extended version): Julia 1.9.2 and R $\geq$ 4.2

License: Inria academic research-only license

Any restrictions to use by non-academics: Yes; prior explicit permission from Inria is required for any non-academic or commercial use (contact stip-sam@inria.fr)

\bibliography{sn-bibliography}

\clearpage
\setcounter{table}{0}
\renewcommand{\tablename}{Supplementary table}
\renewcommand{\thetable}{S\arabic{table}}
\renewcommand{\theHtable}{S\arabic{table}}

\section*{Supplementary Information}

\begin{center}
\small
\begin{longtable}{p{2.5cm}p{9.3cm}p{2cm}}
\caption{Benchmarked FS methods} \\
\toprule
Method & Description & Reference \\
\midrule
\endfirsthead
\multicolumn{3}{c}%
{{\bfseries \tablename\ \thetable{} -- continued from previous page}} \\
\toprule
Method & Description & Reference \\
\midrule
\endhead
\midrule \multicolumn{3}{r}{{Continued on next page}} \\
\endfoot
\bottomrule
\endlastfoot
\multicolumn{3}{c}{\textbf{Embedded methods}} \\
\midrule
LASSO & Logistic regression with the L1 penalty. & \cite{Tibshirani_1996} \\
Adaptive LASSO & LASSO variant that applies adaptive weights to each coefficient, allowing to penalize unimportant variables more than important ones. & \cite{Zou_2006}\\
Exclusive LASSO & Applies the L1 penalty at group level:

$[
\arg\min_{\beta} \; \frac{1}{n} \sum_{i=1}^n \ell(y_i, x_i^\top \beta)
\;+\;
\lambda \sum_{g \in \mathcal{G}} \frac{\|\beta_g\|_1^2}{2}
]$

In \textit{ROOFS}, groups $\mathcal{G}$ are defined via hierarchical clustering, with with $\mid \mathcal{G} \mid = 0.1 \cdot p$ ($p$ is the total number of features in the data).
& \cite{Zhou_Jin_Hoi_2010}\\
\midrule
\multicolumn{3}{c}{\textbf{Ensemble embedded methods}} \\
\midrule
Bolasso & Modified bootstrap LASSO. Selects features with a bootstrap frequency of selection of $\geq 0.5$ instead of 1.& Our modification of \cite{Bach_2008} \\
Stability selection & LASSO-based subsampling method that performs repeated fits across a range of regularization parameters and selects features selected with a frequency above a specified threshold. & \cite{Meinshausen_Buhlmann_2010} \\
Stabl LASSO & Recent extension of stability selection that injects artificial noise variables to enable data-driven choice of the selection frequency threshold. & \cite{Hedou_Maric_2024} \\
RENT & Repeated Elastic Net Technique, a resampling-based method that selects features that satisfy three criteria: a sufficiently high selection frequency across resampled models, consistent coefficient sign across models, and a sufficiently large ratio of the mean coefficient to its standard error. & \cite{rent}\\
Block HSIC LASSO & Method that the Hilbert-Schmidt Independence Criterion (HSIC) to identify non-linear dependencies between features and the outcome while penalizing redundant features. To reduce memory complexity for the case of high-dimensional data, the method partitions samples into smaller blocks. The final bagging block HSIC estimator is computed by averaging HSIC values across multiple random permutations of the data partitions. & \cite{Climente_2019} \\
\midrule
\multicolumn{3}{c}{\textbf{Filter methods}} \\
\midrule
CIFE & Iteratively selects features with the highest Conditional Infomax Feature Extraction (CIFE) score until the user-specified set size is reached. For each feature $X_k$:

$\text{CIFE}(X_k) = I(X_k; Y) - \sum_{X_j \in \mathcal{S}}I(X_j; X_k) + \sum_{X_j \in \mathcal{S}}I(X_j; X_k \mid Y)$

The first term maximizes mutual information (MI) between the feature and the outcome, the second penalizes redundancy with previously selected features, and the third rewards conditional redundancy given the outcome. &  \cite{data_perspective, cife} \\
CMIM & Iteratively selects features with the highest Conditional Mutual Information Maximization (CMIM) score until the user-specified set size is reached. For each feature $X_k$:

$\text{CMIM}(X_k) = \text{min}_{X_j \in \mathcal{S}} [I(X_k; Y \mid X_j)]$

The CMIM score maximizes the minimal conditional MI between the candidate feature $X_k$ and the outcome $Y$, with the minimum value chosen across all previously selected features $X_j \in \mathcal{S}$. & \cite{data_perspective, cmim}  \\
DISR & Iteratively selects features with the highest Double Input Symmetrical Relevance (DISR) score until the user-specified set size is reached. For each feature $X_k$:

$\text{DISR}(X_k) = \sum_{X_j \in F}\frac{I(X_jX_k; Y)}{H(X_jX_kY)}$, where 

$I(X_jX_k; Y) = I(X_k; Y) + I(X_j; Y \mid X_k)$, and 

$H(X_jX_kY) = H(X_k) + H(X_k|X_j) + H(Y|X_k) - I(Y; X_j | X_k)$. & \cite{data_perspective, disr} \\
JMI & Iteratively selects features with the highest Joint Mutual Information (JMI) score until the user-specified set size is reached. For each feature $X_k$: 

$\text{JMI}(X_k) = I(X_k;Y) - \beta \sum_{X_j \in \mathcal{S}} I(X_j; X_k) + \gamma \sum_{X_j \in \mathcal{S}} I(X_j; X_k \mid Y)$, where 

$\beta = \gamma = \frac{1}{|\mathcal{S}|}, \text{where } \mid\mathcal{S}\mid \text{ is the cardinality of a set to be selected} $. & \cite{data_perspective, jmi}  \\
ReliefF & Ranks features by a distance-based score and selects top-k features. For each sample, the algorithm penalizes the features with different values in the nearest neighbors (5 by default) of the same class and rewards the features with different values for the nearest neighbors (similarly, 5 by default) of the opposite class.
&  \cite{data_perspective, relieff} \\
Fisher score &  Ranks features by the ratio of between-class variance to within-class variance and selects top-k features. & \cite{data_perspective} \\
Gini index & Ranks features by their minimum Gini impurity (computed by testing all possible binary splits) and selects top-k features. & \cite{data_perspective} \\
t-score & Ranks features by their two-sample t-statistic and selects top-k features. & \cite{data_perspective} \\
mRMR & Iteratively adds features with the highest relevance-redundancy ratio until the desired number of features is reached. Relevance is measured by the ANOVA F-value, redundancy is iteratively updated as the mean correlation with previously selected features. &  \\
Hierarchical clustering & Performs hierarchical clustering of features based on Spearman correlation coefficients, with the number of clusters equal to the desired number of features to be selected. From each cluster, the feature with the highest point biserial correlation with the outcome is selected.& Our modification of \cite{hierarchical} \\
p.adjust t-test + logistic .05 & Union-based filter that combines features selected by t-test and logistic regression (controlling for user-specified covariates; PD-L1 in the case of PIONeeR data), where p-values from each test were adjusted using the Benjamini-Hochberg procedure with a 0.05 threshold. & Our modification of p-value filtering \\
\midrule
\multicolumn{3}{c}{\textbf{Wrapper methods}} \\
\midrule
LR-RFE & Recursive feature elimination that iteratively removes 10\% of features with the lowest absolute coefficients from a logistic regression model with the L2 penalty until the pre-specified number of features is reached.& \cite{Guyon_2002}\\
RF-RFE & Recursive feature elimination that iteratively removes 10\% of the least important feature of a random forest model and selects a subset with the best AUC in cross-validation. & \cite{Jiang_2004} \\
Forward RF & Recursive forward selection that iteratively adds the most important feature of a random forest model until a pre-specified set size is reached and selects a subset with the best AUC in cross-validation. & \cite{Xia_Yang_2022}\\
Shapicant & Wrapper that uses SHAP importances together with outcome permutation. The criterion of selection is the threshold on p-values from testing whether each feature's SHAP value under the real outcome differs from that under the permuted outcome. In \textit{ROOFS}, the default procedure begins with a threshold of 0.1 and iteratively decreases it, stopping after three iterations or once the number of selected features $k$ falls below the user-specified max\_num\_selected. & \cite{shapicant} \\
\midrule
\multicolumn{3}{c}{\textbf{Controls}} \\
\midrule
Random FS & Selects a feature subset randomly. For the benchmark on PIONeeR data, subset sizes $k$ were randomly sampled from the range 1-20 for the real dataset. For the semi-synthetic dataset, subset size was either set to $k = 19$ (a fixed size corresponding to the number of predictors in the true model) or sampled randomly from [1, $p$]. & \\
Full model & Model trained with all features. & \\
\label{sup_tab1}
\end{longtable}
\end{center}

\subsection*{Choosing an optimal subset size}
\label{sup:size}

For FS methods that select a fixed number of features, we recommend determining the optimal subset size by evaluating each method across a range of sizes using the \texttt{optimize\_num\_selected} function. For each combination of FS method, classifier, and subset size, it computes two metrics $m_1$ and $m_2$ (optimism-corrected AUC and FS stability by default). Then, for each method-classifier pair, an optimal subset size is identified via a distance-based criterion. Both metrics are first min-max normalized, and a weighted Euclidean distance to the point $(1, 1)$ is computed as:

\begin{equation}
    d = \sqrt{w_1\,(1 - m_1)^2 +w_2\,(1 - m_2)^2}
    \label{eq:subset_distance}
\end{equation}

\noindent where by default $m_1$ and $m_2$ are the normalized AUC and stability, respectively, and $w_1 = 0.8, w_2 = 0.2$, but these can be modified by the user. The weighting reflects the higher importance of predictive performance. 

Then, a distance threshold is defined as:

\begin{equation}
    d^* = d_{\min} + 0.5\,\sigma_d
    \label{eq:subset_threshold}
\end{equation}

\noindent where $d_{\min}$ is the minimum observed distance and $\sigma_d$ is its standard  deviation across all tested settings. Among all subset sizes satisfying $d \leq d^*$, the smallest is selected as the recommended value to favor sparser feature sets that do not compromise predictive performance or stability. Results are visualized using a scatter plot, and the recommended configuration is highlighted for each method-classifier combination.

\begin{table}[ht]
\caption{ML classifiers used in FS benchmark on PIONeeR dataset}
\vspace{0.5em}
\centering
\footnotesize
\begin{tabular}{llp{8cm}}
\toprule
Category & Classifier & Hyperparameters\textsuperscript{a} \\
\midrule

\multirow{8}{*}[0pt]{\parbox[c]{1.5cm}{\centering{Tree-based}}}
& BaggingClassifier & \texttt{base\_estimator}=DecisionTree, \texttt{n\_estimators}=200 \\
& & \texttt{max\_depth}=2, \texttt{min\_samples\_leaf}=40, \texttt{max\_features}=0.7 \\
\cmidrule(l){2-3}
& ExtraTreesClassifier & \texttt{n\_estimators}=200, \texttt{max\_depth}=2 \\
& & \texttt{min\_samples\_leaf}=40, \texttt{max\_features}=0.7 \\
\cmidrule(l){2-3}
& RandomForestClassifier & \texttt{n\_estimators}=200, \texttt{max\_depth}=2 \\
& & \texttt{min\_samples\_leaf}=40, \texttt{max\_features}=0.7 \\
\midrule

\multirow{6}{*}[0pt]{\parbox[c]{1.5cm}{\centering{Boosting}}}
& GradientBoostingClassifier & \texttt{max\_depth}=2, \texttt{min\_samples\_leaf}=40 \\
& & \texttt{max\_features}=0.7, \texttt{subsample}=0.7 \\
\cmidrule(l){2-3}
& HistGradientBoostingClassifier & \texttt{max\_depth}=2, \texttt{min\_samples\_leaf}=40, \texttt{max\_features}=0.7 \\
\cmidrule(l){2-3}
& XGBClassifier & \texttt{max\_depth}=2, \texttt{n\_jobs}=-1, other parameters at default \\
\midrule

\multirow{2}{*}[0pt]{\parbox[c]{1.5cm}{\centering{Linear}}}
& LogisticRegression & \texttt{penalty}=None, \texttt{n\_jobs}=-1 \\
\cmidrule(l){2-3}
& LinearDiscriminantAnalysis & Default hyperparameters \\
\midrule

\multirow{4}{*}[0pt]{\parbox[c]{1.5cm}{\centering{Other}} }
& KNeighborsClassifier & Default \\
\cmidrule(l){2-3}
& MLPClassifier & Default \\
\cmidrule(l){2-3}
& SVC & \texttt{probability}=True, other parameters at default \\
\midrule

\multirow{4}{*}[0pt]{\parbox[c]{1.5cm}{\centering{Ensemble}}}
& \multirow{4}{*}[0pt]{\parbox[c]{1.5cm}{\centering{StackingClassifier\textsuperscript{b}}}} & \texttt{cv}=5, \texttt{stack\_method}=\texttt{predict\_proba} \\
& & Base estimators: LogisticRegression, RandomForest, \\
& & SVC, GradientBoosting \\
& & Final estimator: LogisticRegression \\

\bottomrule
\end{tabular}

\begin{flushleft}
\footnotesize
\textsuperscript{a} All classifiers use scikit-learn default parameters unless specified.\\
\textsuperscript{b} StackingClassifier's base estimators use the same hyperparameters as the individual classifiers.
\end{flushleft}
\label{tab:clfs}
\end{table}

\setcounter{figure}{0}
\clearpage
\renewcommand{\figurename}{Supplementary figure}
\renewcommand{\thefigure}{S\arabic{figure}}
\renewcommand{\theHfigure}{S\arabic{figure}}

\begin{figure}[ht]
    \centering
    \includegraphics[width=\textwidth]{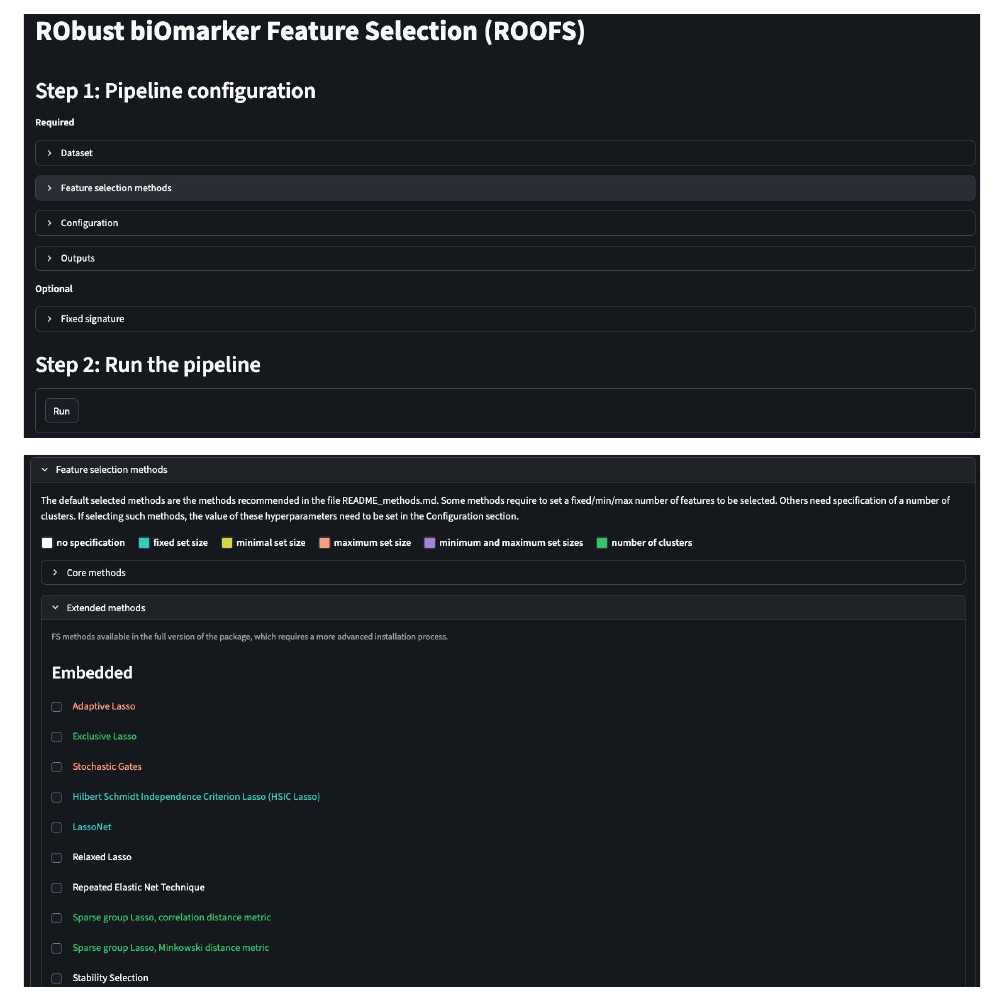}
    \caption{Graphical user interface (GUI) for configuring and executing the \textit{ROOFS} pipeline. In Step 1 (Pipeline configuration), a user should specify the required inputs, including the dataset, chosen FS methods, pipeline parameters, and output settings. Step 2 (Pipeline execution) launches the analysis. The lower panel illustrates the method selection interface. FS methods are organized into core and extended categories depending on the \textit{ROOFS} version. The color coding indicates subtypes based on whether a method selects a signature of fixed or variable size.}
    \label{fig:gui}
\end{figure}

\begin{figure}[ht]
    \centering
    \includegraphics[width=\textwidth]{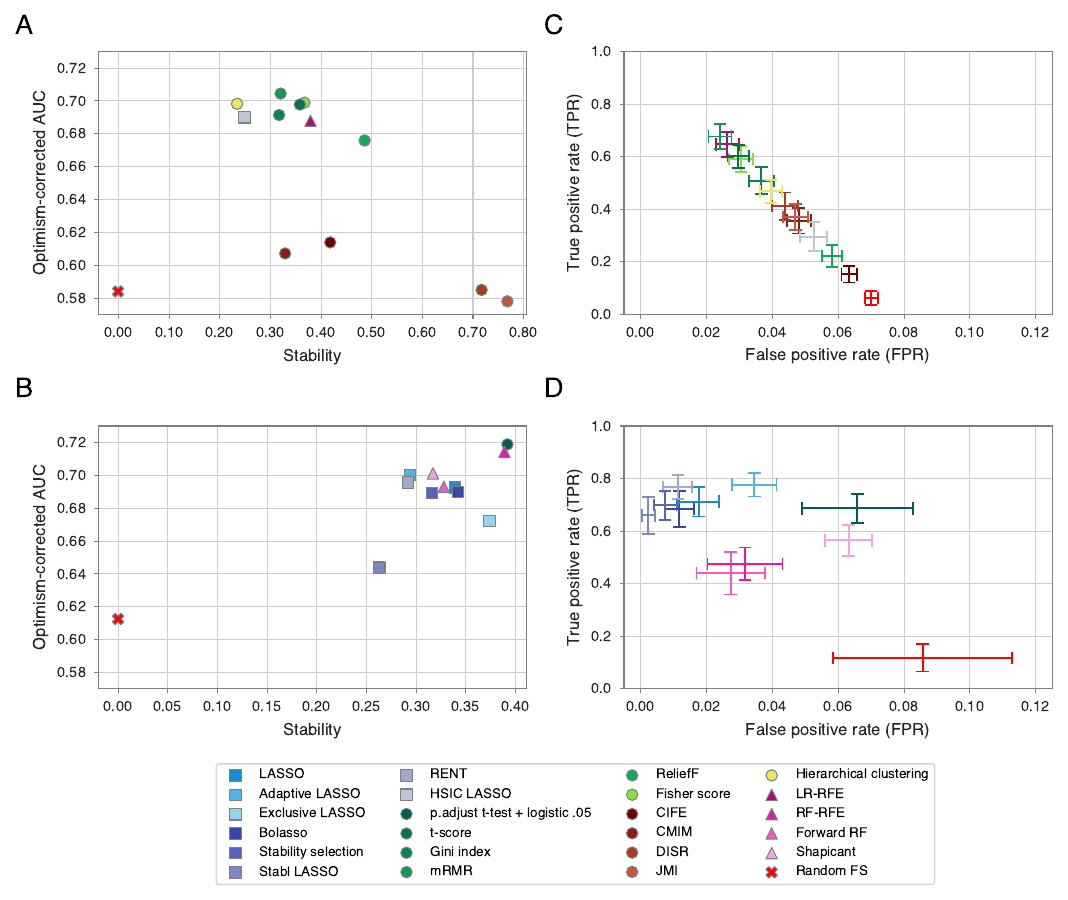}
    \caption{Performance of all benchmarked methods on PIONeeR data (top: FS methods with fixed, user-defined subset size; bottom: FS methods with varying subset size dependent on algorithm thresholds). Blue: embedded methods; violet-blue: resampling-based ensemble methods; green: statistical and distance-based filters; brown: information theory filters; yellow: correlation-based filter; rose: wrappers. \textbf{A, B:} Stability AUC in the experiment on the post-VIF PIONeeR dataset. \textbf{C, D:} Trade-off between discovery and error in the experiment on the semi-synthetic PIONeeR dataset with a simulated outcome. The true positive rate (TPR) shows the proportion of true predictors correctly identified by each FS method, while the false positive rate (FPR) shows the proportion of non-informative features incorrectly selected. Error bars represent one-half of the standard deviation (SD).}
    \label{fig:all-methods}
\end{figure}

\end{document}